\documentclass[lettersize,journal]{IEEEtran}
\usepackage{amsmath,amsfonts}
\usepackage{algorithmic}
\usepackage{algorithm}
\usepackage{array}
\usepackage[caption=false,font=normalsize,labelfont=sf,textfont=sf]{subfig}
\usepackage{textcomp}
\usepackage{stfloats}
\usepackage{url}
\usepackage{verbatim}
\usepackage{graphicx}
\usepackage{cite}
\usepackage{orcidlink}

\usepackage{subfig} 
\usepackage{caption}
\usepackage{booktabs}
\usepackage{lscape}      
\usepackage[justification=raggedright,singlelinecheck=false]{caption}
\usepackage{hyperref}
\renewcommand{\arraystretch}{1.35} 
\usepackage{float}
\makeatletter
\g@addto@macro\scriptsize{
  \setlength\abovedisplayskip{2pt}
  \setlength\belowdisplayskip{2pt}
  \setlength\abovedisplayshortskip{1pt}
  \setlength\belowdisplayshortskip{1pt}
}\normalsize
\g@addto@macro\normalsize{
  \setlength\abovedisplayskip{4pt}
  \setlength\belowdisplayskip{4pt}
  \setlength\abovedisplayshortskip{2pt}
  \setlength\belowdisplayshortskip{2pt}
}
\makeatother

\usepackage{etoolbox}

\begin{document}
\title{BLIP-FusePPO: A Vision-Language Deep Reinforcement Learning Framework for Lane Keeping in Autonomous Vehicles}

\author{
    Seyed Ahmad Hosseini Miangoleh\textsuperscript{\scalebox{1.25}{\orcidlink{0009-0007-6572-6688}} 1},  
    Amin Jalal Aghdasian\textsuperscript{\scalebox{1.25}{\orcidlink{0009-0003-8482-1219}} 1}, 
    Farzaneh Abdollahi\textsuperscript{\scalebox{1.25}{\orcidlink{0000-0003-4957-987X}} 1}\\
    \textsuperscript{1}Department of Electrical Engineering, Amirkabir University of Technology (Tehran Polytechnic), Tehran, Iran
}
\maketitle
\begin{abstract}
In this paper, we propose Bootstrapped Language–Image Pretraining-driven Fused State Representation in Proximal Policy Optimization (BLIP-FusePPO), a novel multimodal reinforcement learning (RL) framework for autonomous lane-keeping (LK), in which semantic embeddings generated by a vision-language model (VLM) are directly fused with geometric states, LiDAR observations, and Proportional-Integral-Derivative-based (PID) control feedback within the agent observation space. The proposed method lets the agent learn driving rules that are aware of their surroundings and easy to understand by combining high-level scene understanding from the VLM with low-level control and spatial signals. Our architecture brings together semantic, geometric, and control-aware representations to make policy learning more robust. A hybrid reward function that includes semantic alignment, LK accuracy, obstacle avoidance, and speed regulation helps learning to be more efficient and generalizable. Our method is different from the approaches that only use semantic models to shape rewards. Instead, it directly embeds semantic features into the state representation. This cuts down on expensive runtime inference and makes sure that semantic guidance is always available. The simulation results show that the proposed model is better at LK stability and adaptability than the best vision-based and multimodal RL baselines in a wide range of difficult driving situations.We make our code publicly available.\footnote{\href{https://github.com/Amin-A96/BLIP-FusePPO-A-Vision-Language-Deep-Reinforcement-Learning-Framework-for-Lane-Keeping-in-Autonomous.git}{GitHub Repository}}

\end{abstract}


\section{Introduction}
\IEEEPARstart{A}{utonomous} vehicle (AV) technology has expanded enormously from a futurist theory to a realistic solution for modern transportation problems over the past few years. Even if Advanced Driver Assistance Systems (ADAS) have greatly enhanced safety and efficiency, inefficiencies, high accident rates, and non-optimal traffic flow still prevail.In this regard, LK systems play a vital role as essential components of Level 1 and Level 2 autonomy \cite{liu2021vision}, although they often face limitations in real-world environments. Driven by intelligent perception and decision-making systems, AVs present a feasible future by reading and reacting to dynamic, complex environments in real-time \cite{pan2024impacts}. RL is emerging as a strong framework among several machine-learning methods for training autonomous agents through continuous interaction and adaptation. Semantic context has recently demonstrated the remarkable ability of VLMs to enhance visual perception through spatial reasoning. More robust  driving rules are generated by mixing VLMs with RL, so enabling safer and more effective autonomous navigation \cite{hoel2019combining,getahun2024integrated}.

\subsection{Related Work}

Classic LK systems, utilizing modular designs, integrate classical computer vision methods, including edge detection and Hough transforms, with control strategies such as PID or Model Predictive Control (MPC) controllers \cite{krishnan2024real,wen2023distributed}. These systems are efficient under ideal conditions. However they don't perform well in real-world environments, including worn lane marks, different illumination, or blocked lanes. Their adaptability is limited and requires significant scenario-specific tuning to ensure strong performance. They also largely depend on precise modeling of the vehicle's dynamics \cite{lee2025lane}. Using professional demonstrations to acquire desired behaviors provides imitation learning with a strong basis for training autonomous systems \cite{le2022survey}. Its main advantage is its efficiency, in avoiding the need for explicitly codifying complex reward functions.
Furthermore, achieving fast-reaching human-like performance in structured environments with plentiful and high-quality expert data is facilitated by imitation learning \cite{teng2022hierarchical} and behavioral cloning (BC) \cite{bhattacharyya2022modeling}. Still, imitation learning does not come without challenges. Its dependence on the quality and diversity of the demonstration data is minimal; models sometimes struggle to extend to unseen events or edge cases not addressed in the training set. This can lead to compounding errors, especially in dynamic or unpredictable environments since even slight deviations from the expert policy magnify over time \cite{kuutti2020survey}.

With great success in LK applications, RL, especially in its deep reinforcement learning (DRL) form, has become a potent framework for developing adaptive control policies in autonomous driving \cite{he2023deep}. DRL enables agents to learn visuomotor behaviors through interaction with dynamic and uncertain environments by utilizing neural networks to approximate complex control functions \cite{abdollahian2024enhancing}. Early discoveries, such as the Deep Q-Network (DQN), demonstrated the efficacy of combining convolutional neural networks (CNNs) with experience replay for discrete control tasks, thereby laying the basis for vision-based driving \cite{sharma2024reinforcement}. Using parallelized policy updates, actor-critic approaches such as Asynchronous Advantage Actor-Critic (A3C) \cite{zhou2023lane} extended this capability to continuous domains. More recently developed algorithms, such as PPO \cite{wu2023dyna}, have garnered significant interest due to their stability and efficiency in continuous control settings. These qualities are essential for tasks including steering and speed control. Using deterministic and entropy-regularized stochastic policies, respectively, other techniques such as Deep Deterministic Policy Gradient (DDPG) and Soft Actor-Critic (SAC) further improve sample efficiency in continuous action environments \cite{perez2022deep,aghdasian2023autonomous}.

Unlike value-based methods such as DQN, which are limited to discrete actions and not proper fine-grained control, PPO directly optimizes a stochastic policy using clipped surrogate objectives that promote stable and monotonic policy updates. PPO provides consistent convergence and reduced gradient variance compared to A3C. Although DDPG and SAC are sensitive to initialization and noise and require significant hyperparameter tuning \cite{cheng2023lane}, they perform relatively well in continuous domains. On the other hand, PPO is a recommended choice for real-world autonomous driving tasks, as it strikes a reasonable balance between performance and training stability.

In \cite{ge2024llm} shows the integration of large language models (LLMs) into intelligent operating systems (LLM-OSs) to play a role as contextual reasoning for AVs, which offers a unified architecture but introduces new implementation challenges. In \cite{cui2023drivellm}, a decision-making framework is proposed that integrates LLM into AD to enable understanding reasoning, interactive human instruction handling, and cyber-physical feedback for improved performance in complex and adversarial scenarios. VLMs have become an efficient method \cite{alsabbagh2025minimedgpt,chen2024advanced} in automated driving by providing a mutual connection between visual observations and natural language semantics.  Strong semantic representations provided by VLMs enable autonomous agents to understand complex environments linguistically grounded in visual perception. Object referencing, scene captioning, open-vocabulary object detection, and visual question answering (VQA) \cite{tian2024drivevlm,sima2024drivelm} are among the broad spectrum of tasks relevant to autonomous driving that VLMs support by bridging computer vision and natural language processing. Safe and dependable deployment in real-world driving conditions depends on interpretability and human-machine interaction, which this semantic integration improves \cite{malla2023drama}.

Recent research uses large-scale contrastive and generative pretraining to improve generalization in self-driving cars. This is based on earlier work with foundational VLMs, such as Contrastive Language–Image Pretraining (CLIP) and  Bootstrapping language-image pre-training (BLIP). CLIP utilizes image-text contrastive learning to create a shared semantic space, enabling it to recognize rare road elements without any prior training. BLIP adds vision-language understanding and generation, which enables models to describe scenes and infer what someone wants to do. These features make tasks such as VQA, open-vocabulary detection, and following instructions, making autonomous systems be safer and easier to understand.\cite{you2022learning}.

Five main domains comprise a structured taxonomy of VLM applications in autonomous driving: perception, navigation, planning, decision-making, and data generating. Models, including TransRMOT and PromptTrack, use language-conditioned tracking and multi-view localization in the perception domain. Systems like ALT-Pilot navigate using linguistic markers matched with topometric maps to carry out commands like "turn left after the traffic light." Driven by natural language to define scene context, estimate risk, and guide control, recent open-loop reasoning models, including DriveGPT4 and DLaH, show the value of VLMs in complex decision-making. More recently, VLMs have been combined with RL \cite{venuto2024code} to address the challenge of learning driving policies from high-level language goals rather than manually engineered reward functions. In autonomous driving, where conventional reward design is labor-intensive and often fails to generalize to various traffic environments, this is especially valuable. VLM-RL employs a batch-processing mechanism with a replay buffer to compute rewards asynchronously, thereby significantly reducing inference overhead during training and ensuring scalability.

\subsection{Motivations of the Paper}

The majority of vision-based RL methods for autonomous LK use either raw visual features or shallow visual cues alone. Such models tend to miss the semantic context of the driving scene and generalize poorly to complicated or unseen situations, e.g., occluded lanes or ambiguous road markings. Furthermore, current multimodal RL methods only employ VLMs for reward shaping instead of injecting them into the agent observation space. Consequently, the agent is deprived of the capacity to reason about both semantic and control domains while learning. This serves as the first primary motivation for this work.

In addition, conventional RL-based controllers are not interpretable and robust, especially under dynamic or uncertain conditions. If without access to structured control feedback, their behaviors could be unstable or unexplainable. Supplementing the observation space with interpretable control features, lateral deviation, heading angle, and velocity error from PID controller can improve both training stability and policy reliability. This is the second motivation of this paper.

Finally, VLM-driven RL systems that calculate semantic rewards during runtime are plagued with heavy computational overheads, which cause higher inference latency and hinder real-time deployment. Skirting this overhead by encoding semantic information within the state space itself enables the system to maintain semantic awareness without compromising efficiency. This constitutes the third major motivation for this paper.

\subsection{Contributions of the Paper}
To bridge the gap between semantic perception and control-aware policy learning in autonomous lane keeping, this work proposes a novel architecture that integrates semantic features extracted by BLIP and control feedback from a classical PID controller directly into the state representation used by the reinforcement learning agent. Furthermore, it introduces a hybrid reward formulation that combines both semantic and geometric cues to support efficient and robust policy learning. The main contributions of this work are:

\begin{itemize}
    \item \textbf{A novel architecture} that unifies semantic perception from BLIP with policy learning that is aware of control, by injecting language-conditioned embeddings and PID control signals directly into the reinforcement learning agent's state representation.

    \item \textbf{A state augmentation technique based on PID control} that enriches the observation space with interpretable control errors such as lateral deviation, heading angle, and speed offset. This enhancement improves learning stability and policy robustness when using PPO.

    \item \textbf{The design of a novel hybrid reward function} that integrates semantic alignment from BLIP, geometric lane adherence, and velocity regularization to accelerate convergence and enhance generalization across diverse driving scenarios.
\end{itemize}

The structure of this paper is organized as follows: Section~\ref{sec:preliminaries} provides the preliminaries and problem formulation. Section~\ref{sec:methodology} details the proposed BLIP-FusePPO methodology and hybrid state representation. Section~\ref{sec:Results} presents the simulations and results. Finally, Section~\ref{sec:Conclusion} discusses the conclusions of the paper.

\section{Preliminaries}
\label{sec:preliminaries}

\subsection{Partially Observable Markov Decision Process (POMDP)}

The AV LK task is set up as a POMDP, which is shown by the tuple $(\mathcal{S}, \mathcal{A}, \mathcal{T}, \mathcal{R}, \mathcal{O}, \mathcal{Z}, \gamma)$. The parts are defined like this:

\begin{itemize} \item \textbf{State Space} ($\mathcal{S}$): The complete, hidden picture of the environment, including the vehicle position, the road and lane geometry, nearby obstacles, and the changing context. The agent cannot see this state directly.

     \item \textbf{Action Space} ($\mathcal{A}$): A continuous control domain where the agent makes steering and speed commands to keep the car in the lane and moving forward.
    
     \item \textbf{Transition Function} ($\mathcal{T}$): A stochastic function $\mathcal{T}(s_{t+1} \mid s_t, a_t)$ that shows how the state of the environment changes when the agent does something, taking into account the vehicle movement and how it interacts with the environment.
    
     \item \textbf{Reward Function} ($\mathcal{R}$): A scalar feedback function $\mathcal{R}(s_t, a_t)$ that rewards good behavior, like staying in the lane, avoiding obstacles, and keeping a safe speed, and punishes bad behavior, like making unsafe or unstable moves.
    
     \item \textbf{Observation Space} ($\mathcal{O}$): A multimodal vector that includes sensory measurements like visual input, range data, control feedback signals, and semantic descriptors that were taken from perceptual observations.
    
     \item \textbf{Observation Function} ($\mathcal{Z}$): A probabilistic mapping $\mathcal{Z}(o_t \mid s_t)$ that shows how the agent noisy, partial observations are related to the actual state of the environment, taking into account sensor limitations and uncertainty in perception.
    
     \item \textbf{Discount Factor} ($\gamma$): A scalar $\gamma$ that ranges from 0 to 1 and controls the balance between short-term and long-term rewards in the cumulative return objective.
\end{itemize} 

The agent goal is to find a policy $\pi(a_t \mid o_t)$ that maximizes the expected cumulative discounted reward while working with incomplete sensory data and not knowing what will happen next. This POMDP formulation captures the main problems that self-driving cars face in the real world: not capable of seeing everything, have to handle with different types of information, and changing situations. 

\subsection{BLIP}
\label{sec:vlms}

Here, we utilize BLIP\cite{cho2025enhanced} as the VLM module to obtain semantically rich text descriptions from raw visual inputs, enabling us to abstract and generalize beyond pixel-level features.

The front-facing RGB camera takes a picture of the driving scene at each discrete time step. The BLIP encoder then encodes this picture. The BLIP model generates a text description, called $c_t$, that conveys important aspects of the scene, such as "a two-lane curved road with a car ahead" or "lane markings with no obstacles in sight." This description gives you important semantic information that you need to make smart choices while driving.

The language encoder in the BLIP model tokenizes the caption $c_t$ and then puts it into a fixed-dimensional vector $e_t \in \mathbb{R}^d$. The embedding adds to the lower-level sensor data by giving a short semantic summary of the visual scene. 

\subsection{Problem Statement}

The goal of this work is to learn an optimal policy $\pi: \mathcal{O} \rightarrow \mathcal{A}$ that maximizes the expected cumulative reward:

\begin{equation}
\pi^* = \arg\max_{\pi} \mathbb{E}_{\pi} \left[ \sum_{t=0}^{\infty} \gamma^t \mathcal{R}(s_t, a_t) \right]
\label{eq:optimal_policy}
\end{equation}

Under the constraints of partial observability and sensor noise. The agent observation at each timestep integrates the following modalities:
\begin{enumerate}
    \item Visual input from a front-facing RGB camera,
    \item LiDAR-based range measurements,
    \item Lateral deviation feedback from a classical controller,
    \item Semantic embeddings generated by a pre-trained VLM.
\end{enumerate}

The key challenge is to develop a policy that attains precise and invariant lane navigation. furthermore, generalize well to new environments, various road conditions, and different light or weather conditions. The policy should be robust, sample-efficient, and proficient in utilizing multimodal data for making stable decisions in real-time, uncertain, and dynamic environments.

\section{Methodology}
\label{sec:methodology}

\subsection{Hybrid State Representation}

\begin{figure*}[t]
    \centering
\includegraphics[scale=0.38]{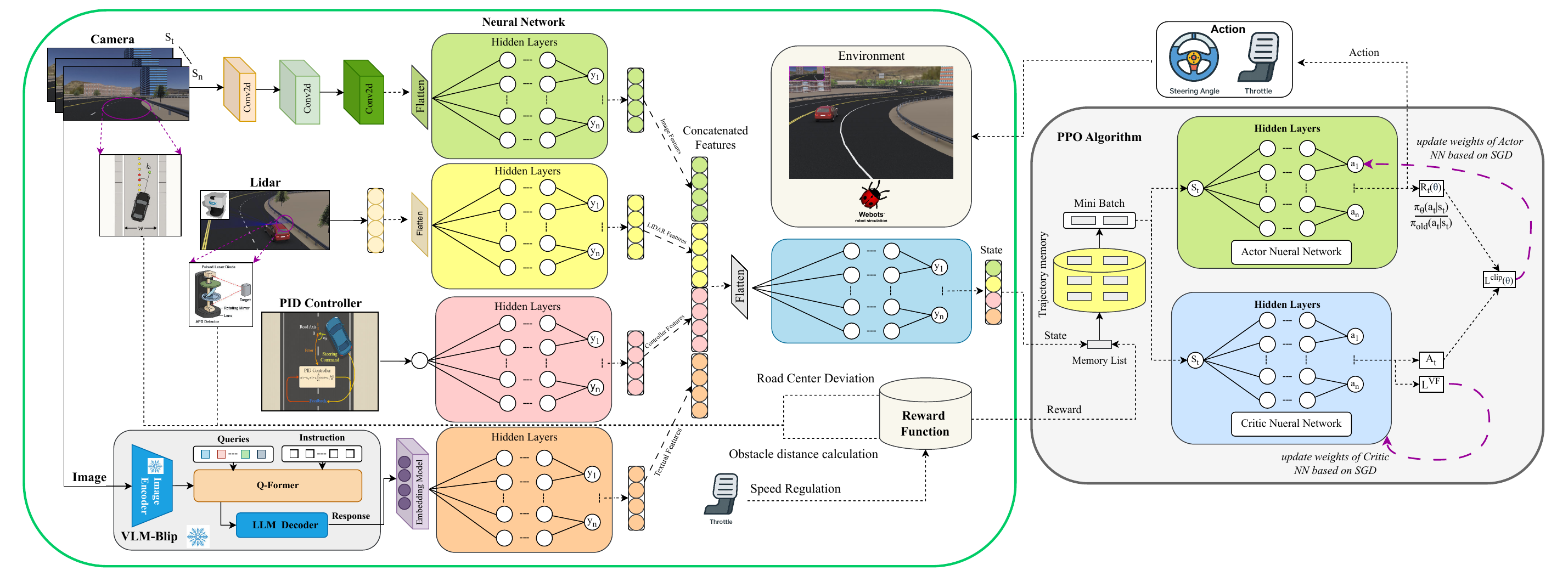}
    \caption{A brief look at the BLIP-FusePPO pipeline. The agent receives data from cameras, LiDAR, PID error feedback, and semantic embeddings from a BLIP. These are processed and combined into a hybrid state vector, which is then passed to a policy trained using PPO.}
    \label{fig:overview}
\end{figure*}

To enable context-aware and safe autonomous LK, we propose BLIP-FusePPO, an RL framework that fuses complementary sensory and control features into a hybrid state representation. As shown in Fig.~\ref{fig:overview}, the agent observation vector integrates four core modalities: (i) RGB visual input from a front-facing camera, (ii) spatial distance measurements from a LiDAR sensor, (iii) lateral correction values computed by a classic PID controller, and (iv) semantic embeddings derived from a VLM. We choose each modality based on how well it adds to perceptual richness, how well it works in different environments, and how well it helps with sample-efficient policy learning. Each component goes through preprocessing and normalization to make sure the scale is consistent, the noise is reduced, and the downstream fusion in the PPO learning pipeline works well.

\subsection*{\textbf{State Components and Preprocessing}}

\textbf{1) Camera Image:}  
A front-facing RGB camera displays the lanes. The lower road area is resized to 224 x 244 pixels, and the pixel values are set to the range [0, 1]. The Hough Transform detects lane markings and calculates the distance from the center, which determines the reward: minor deviations are rewarded, while larger ones are increasingly penalized, encouraging driving closer to the center.

\vspace{1em}
\textbf{2) LiDAR Data:}  
LiDAR provides a 180-degree horizontal range vector, offering extensive spatial awareness without altering the lighting. Values are clipped and normalized, and the closest distance is used to guess how close you are to an obstacle. A segmented reward punishes unsafe distances and rewards safe spacing. This encourages drivers to be aware of obstacles and helps them stay in their lanes.

\vspace{1em}
\textbf{3) PID Correction:}  
The scalar correction output of a classic PID controller is incorporated as an auxiliary state variable. The PID controller estimates the required lateral steering adjustment based solely on the lateral error from the road center. It is defined as:

\begin{equation}
    u_t = K_p e_t + K_i \sum_{\tau=0}^{t} e_\tau \Delta t + K_d \frac{e_t - e_{t-1}}{\Delta t}
\end{equation}

Where $e_t$ is the current lateral error, and $K_p$, $K_i$, $K_d$ are proportional, integral, and derivative gains, respectively. To maintain numerical stability and prevent phenomena such as integral windup or control signal saturation, the integral and control outputs are clipped to remain within predefined bounds. Specifically, the clipping operation constrains both $I_t$ and $u_t$ to specified intervals, ensuring that error accumulation and control commands do not exceed safe or physically meaningful limits: 

\begin{equation} 
	I_t = \text{clip}(I_{t-1} + e_t \Delta t, -I_{\max}, I_{\max}) 
\end{equation} 

\begin{equation} 
	u_t = \text{clip}(u_t, -u_{\max}, u_{\max}) 
\end{equation}

The PID signal is \textit{not} directly used for actuation but serves as a learned feature in the observation vector to support convergence and provide prior control knowledge during policy learning .

\begin{figure*}[t]
    \centering
    \includegraphics[scale=0.75]{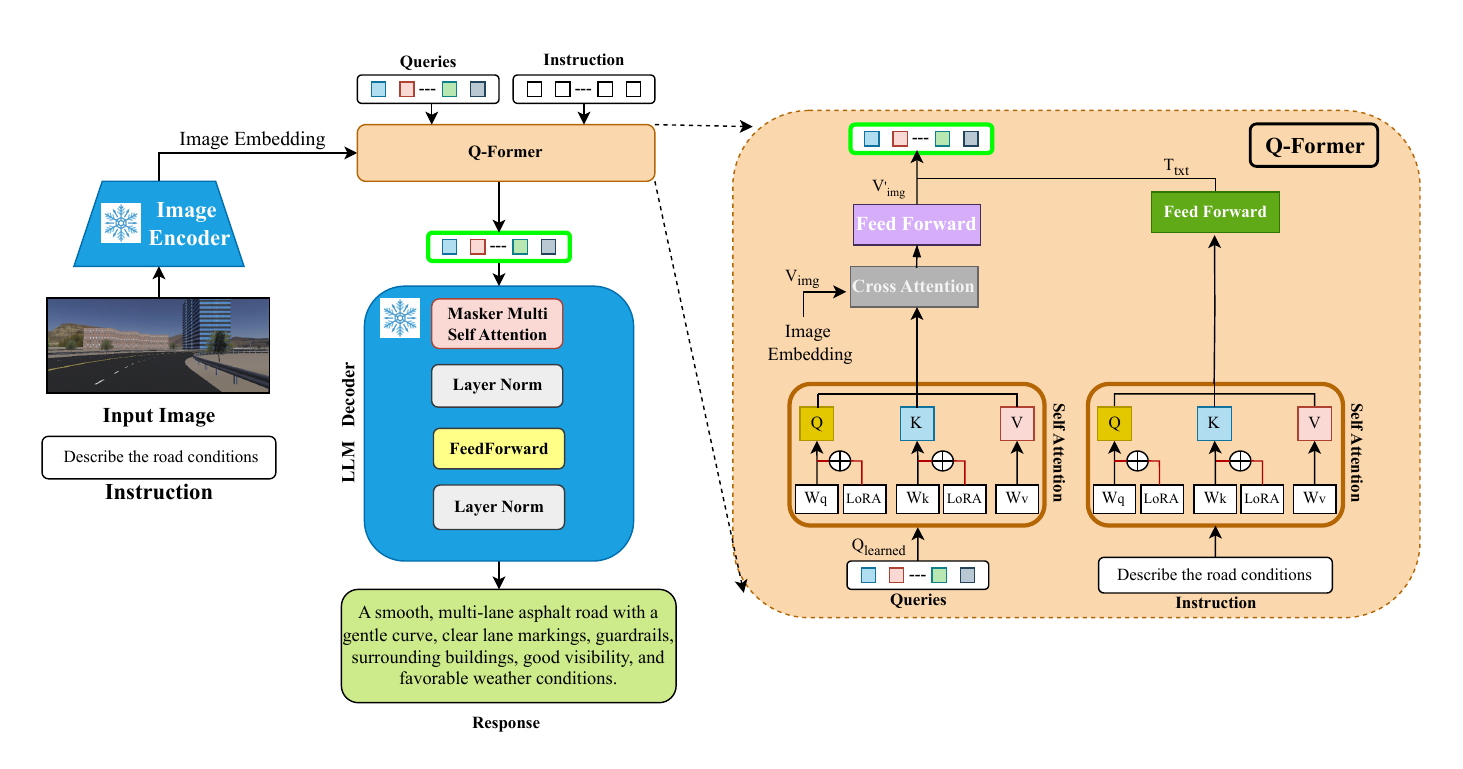}
    \caption{Pipeline for semantic token generation via VLM. Captions are generated and tokenized into fixed-length embeddings.}
    \label{fig:blip_pipeline}
\end{figure*}
\vspace{1em}
\textbf{4) Semantic Tokens from VLM:}\\
BLIP, a pre-trained VLM, processes each RGB image to get high-level semantic context. Fig. \ref{fig:blip_pipeline} shows that BLIP makes a text caption that describes the scene (like "curved road with lane markings") and then breaks it up into a fixed-length vector. The embedding layer adds this vector to the hybrid state vector.

Semantic embeddings offer several key benefits:
\begin{itemize}
	\vspace{0.8em}
    \item \textbf{Abstract understanding:} Capture high-level information about road topology and scene structure.
    \item \textbf{Robustness to visual noise:} Remain stable under pixel-level perturbations such as lighting changes, occlusion, or adverse weather.
    \item \textbf{Improved generalization:} Facilitate policy transfer across unseen environments with diverse appearances.
\end{itemize}

The driving environment scene (\emph{input image}) is first transformed into vision features using a pre-trained Vision Transformer (ViT):
\begin{equation}
    \mathbf{V}_{\text{img}} = \text{ImageEncoder}(\mathbf{x})
    \label{eq:vision-encoder}
\end{equation}
where $\mathbf{x}$ is the input image.

The learned queries $\mathbf{Q}_{\text{learned}}$ are then processed by a self-attention module augmented with LoRA-efficient low-rank adaptation~\cite{cho2025enhanced}:
\begin{equation}
    \mathbf{Q}'_{\text{learned}} = \mathrm{Self Attention}_{\mathrm{LoRA}}(\mathbf{Q}_{\text{learned}})
    \label{eq:self-attn-lora}
\end{equation}
Inside the self-attention mechanism, LoRA is incorporated by applying low-rank matrices to the query and key matrices as follows:
\begin{equation}
\begin{split}
\mathrm{Attention}(\mathbf{Q},\, \mathbf{K},\, \mathbf{V}) = 
\mathrm{softmax}\left(
    \frac{
        \mathbf{Q}_\ell \, \mathbf{K}_\ell^\top
    }{
        \sqrt{d_k}
    }
\right) \mathbf{V}
\end{split}
\label{eq:lora-attention}
\end{equation}
Here, the low-rank adapted query and key projections are defined as:
\[ \mathbf{Q}_\ell = \mathbf{X}W_Q + B_Q A_Q, \quad \mathbf{K}_\ell = \mathbf{X}W_K + B_K A_K \]
where:
\begin{itemize}
    \item \( \mathbf{X} \) is the input feature matrix,
    \item \( W_Q, W_K \) are the original projection weights,
    \item \( A_Q, A_K \in \mathbb{R}^{r \times d} \) and \( B_Q, B_K \in \mathbb{R}^{d \times r} \) are learnable low-rank adaptation matrices,
    \item \( r \ll d \), where \( d \) is the feature dimension and \( r \) is the LoRA rank.
\end{itemize}

where $B_Q = W_{Q,\text{down}} W_{Q,\text{up}}$ and $B_K = W_{K,\text{down}} W_{K,\text{up}}$. 

\begin{itemize}
    \item $\mathbf{X}$ denotes the input features to the self-attention (here, the queries).
    \item $d_k$ is the dimension of the key vectors, used for variance stabilization.
    \item $W_{*,\text{down}} \in \mathbb{R}^{d \times r}$ and $W_{*,\text{up}} \in \mathbb{R}^{r \times d}$ are the LoRA down- and up-projection matrices, with $r \ll d$.
\end{itemize}

The refined queries $\mathbf{Q}'_{\text{learned}}$ and the extracted visual features $\mathbf{V}_{\text{img}}$ undergo cross-attention:
\begin{equation}
    \mathbf{V}'_{\text{img}} = \mathrm{Cross Attention}\left(\mathbf{Q}'_{\text{learned}},\, \mathbf{V}_{\text{img}}\right)
    \label{eq:cross-attn}
\end{equation}

Similarly, the instruction input text is also processed via a LoRA-augmented self-attention layer:
\begin{equation}
    \mathbf{T}_{\text{txt}} = \mathrm{Self Attention}_{\mathrm{LoRA}}(\mathrm{InputText})
    \label{eq:txt-attn}
\end{equation}

Finally, the outputs $\mathbf{V}'_{\text{img}}$ and $\mathbf{T}_{\text{txt}}$ are provided to the LLM decoder, which generates a contextually rich textual caption describing the driving scene:
\begin{equation}
    \hat{y} = \mathrm{LLM\_Decoder}\left([\mathbf{T}_{\text{txt}}; \mathbf{V}'_{\text{img}}]\right)
    \label{eq:llm-decoder}
\end{equation}

\begin{figure}[H]
    \centering
    \includegraphics[scale=0.45]{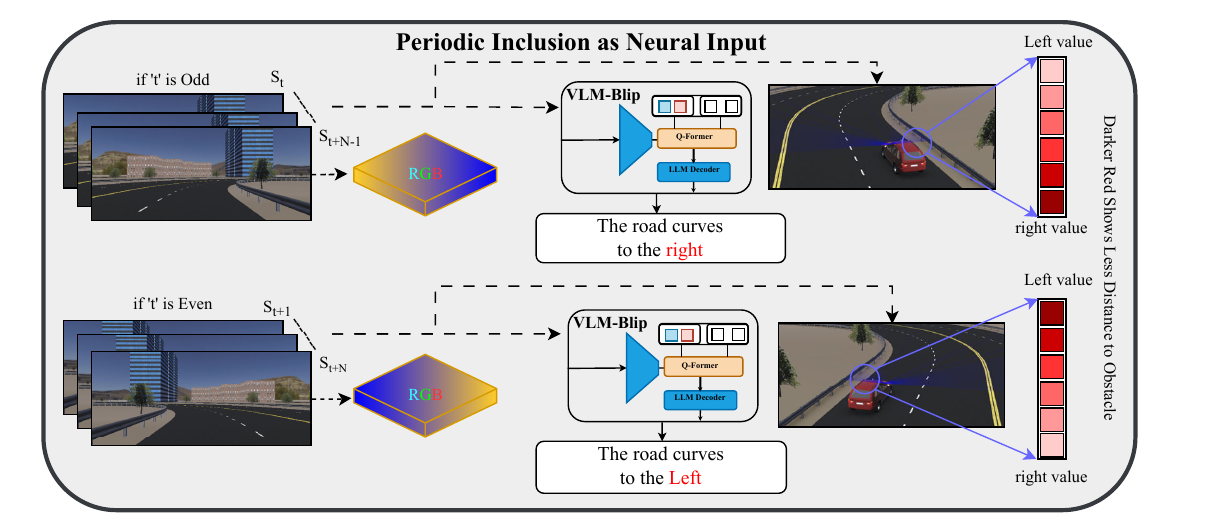}
    \caption{Symmetric augmentation improves generalization and helps mitigate directional bias in training data.}
    \label{fig:aug}
\end{figure}

\vspace{1em}
\textbf{5) Augmentation Strategy:}  
To improve generalization, data augmentation is periodically applied (Fig.~\ref{fig:aug}). Every $T$ timesteps, the state is horizontally mirrored:  
The image is flipped,  
LiDAR readings are reversed,  
PID correction sign is inverted,  
Semantic tokens are regenerated for the flipped image,  
The steering action is mirrored.

This symmetry-aware augmentation improves policy robustness to diverse road geometries and avoids overfitting to dataset-specific layouts.

\subsection*{\textbf{Integration and Feature Fusion}}
Each modality is processed via a dedicated neural branch to maximize its representational fidelity:  
The RGB image is fed into a CNN,  
The LiDAR vector and semantic tokens are processed by separate fully connected (FC) layers,  
The PID correction is embedded via a lightweight FC layer.

The outputs of all branches are concatenated and passed through a fully connected layer with 128 neurons activated by the ReLU function. This results in a compact yet expressive state vector used by both the policy and value networks.

All parts go through normalization or capping to make sure that the numbers stay stable:

- Image pixels are in the range [0,1],  

- The values for LiDAR and PID are clipped and scaled. 

- The semantic tokens are made the same.

This design makes a strong, multimodal state encoding that helps with fast learning, stable convergence, and strong transferability to the real world.

\subsection{Action Space}

The proposed RL framework employs a continuous, two-dimensional action space defined as $\mathbf{a} = (a_1, a_2)$, where each element lies within the normalized interval $[-1, 1]$. This design provides the agent with fine-grained control over both steering and velocity:

\begin{itemize}
    \item \textbf{Steering Angle ($a_1$):} Represents the normalized lateral control command. A value of $a_1 = -1$ maps to the maximum left steering angle, and $a_1 = +1$ corresponds to the maximum right. The physical steering command is computed as:
    \begin{equation}
        \text{Steering} = a_1 \times \text{max\_steering\_angle}
    \end{equation}
    where $\text{max\_steering\_angle}$ denotes the system-specific steering limit.
    
    \item \textbf{Target Speed ($a_2$):} Governs the longitudinal velocity. A value of $a_2 = -1$ denotes a full stop, while $a_2 = +1$ maps to the maximum allowed cruising speed. The conversion to physical units is defined as:
    \begin{equation}
        \text{Target Speed} = \frac{a_2 + 1}{2} \cdot v_{\text{max}}
    \end{equation}
    where $v_{\text{max}}$ denotes the upper speed limit in the simulation
\end{itemize}

During training, both action dimensions are taken from separate Gaussian policies and run directly in the simulator at every time step. This continuous action formulation makes it possible to have smooth, high-resolution control responses that are necessary for driving in the real world. Moreover, separating lateral and longitudinal control makes it learn more flexible and efficient behavior from observations in more than one mode. Normalization makes it easier to learn policies and apply them in both simulations and real-life situations.
\subsection{Reward Function}
\label{sec:reward}

The reward function $R_t$ combines four main goals to encourage safe, centered, and smooth driving: (i) keeping the car from drifting too far to the side, (ii) making sure that LiDAR sensing can clear obstacles, (iii) keeping the car at a target cruising speed, and (iv) giving harsher punishments for big position changes. At timestep $t$, the reward is given as:

\begin{equation}
R_t = w_1 \cdot r_{\text{lane}} + w_2 \cdot r_{\text{lidar}} + w_3 \cdot r_{\text{speed}} + w_4 \cdot r_{\text{center}}
\end{equation}

where $w_i$ are scalar weights to balance the contribution of each reward term.

\subsubsection*{1) LK Reward ($r_{\text{lane}}$)}
Encourages minimal lateral offset $\Delta x$ from the centerline:
\begin{equation}
r_{\text{lane}} = 1 - \frac{|\Delta x|}{d_{\text{lane}}}
\end{equation}
where $d_{\text{lane}}$ denotes a normalization constant in pixels. This reward linearly decreases with increasing deviation.

\subsubsection*{2) Obstacle Avoidance via LiDAR ($r_{\text{lidar}}$)}
Promotes safe spacing from nearby objects based on the minimum LiDAR reading $d_{\min}$. The reward is piecewise defined as:
\begin{equation}
r_{\text{lidar}} =
\begin{cases}
-b_1 \cdot \frac{(d_{\text{mid}} - d_{\min})}{d_{\text{range}}}, & \text{if } d_{\min} \in [d_{\text{low}}, d_{\text{mid}}] \\
-b_2 + b_3 \cdot d_{\min}, & \text{if } d_{\min} < d_{\text{crit}} \\
+r_{\text{bonus}}, & \text{if } d_{\min} \in [d_1, d_2] \cup [d_3, d_4] \\
0 & \text{otherwise}
\end{cases}
\end{equation}

This formulation penalizes proximity violations and provides sparse positive feedback in safe distance ranges.

\subsubsection*{3) Speed Matching Reward ($r_{\text{speed}}$)}
Ensures non-zero cruising by penalizing deviations from a target velocity $v_{\text{target}}$:
\begin{equation}
r_{\text{speed}} = - \left( \frac{v - v_{\text{target}}}{v_{\text{target}}} \right)^2
\end{equation}

which applies a quadratic penalty for deviation above or below the target.

\subsubsection*{4) Centralization Penalty ($r_{\text{center}}$)}
Applies stronger penalties for larger lateral offsets:

\begin{equation}
r_{\text{center}} = -k \cdot \left( \frac{|\Delta x|}{d_{\text{clip}}} \right)^2
\end{equation}
where $k$ is a gain and $d_{\text{clip}}$ defines a saturation threshold.

\subsubsection*{5) Episode Termination Conditions}
Episodes are forcefully terminated and penalized under the following conditions:
\begin{itemize}
    \item $|\Delta x| > d_{\text{reset}}$: the vehicle has exited the road area.
    \item $d_{\min} < d_{\text{fail}}$: repeated violations of safe obstacle distance.
\end{itemize}

A fixed penalty is applied upon early termination:
\begin{equation}
R_t = -r_{\text{fail}}
\end{equation}

\subsubsection*{6) Reward Clipping}
To stabilize learning, all rewards are clipped within a bounded range:
\begin{equation}
R_t \leftarrow \mathrm{clip}(R_t, -r_{\text{clip}}, +r_{\text{clip}})
\end{equation}
\vspace{-0.5em}

To train the LK policy in a high-dimensional, multi-modal observation space, we employ PPO as the RL algorithm. PPO is a first-order, on-policy actor-critic method that combines the benefits of policy gradient approaches and trust region optimization. It achieves both stability and sample efficiency by relying on a clipped surrogate objective, which restricts the magnitude of policy updates, thus preventing destabilizing policy oscillations.

\begin{figure}[H]
    \centering
    \includegraphics[scale=0.3]{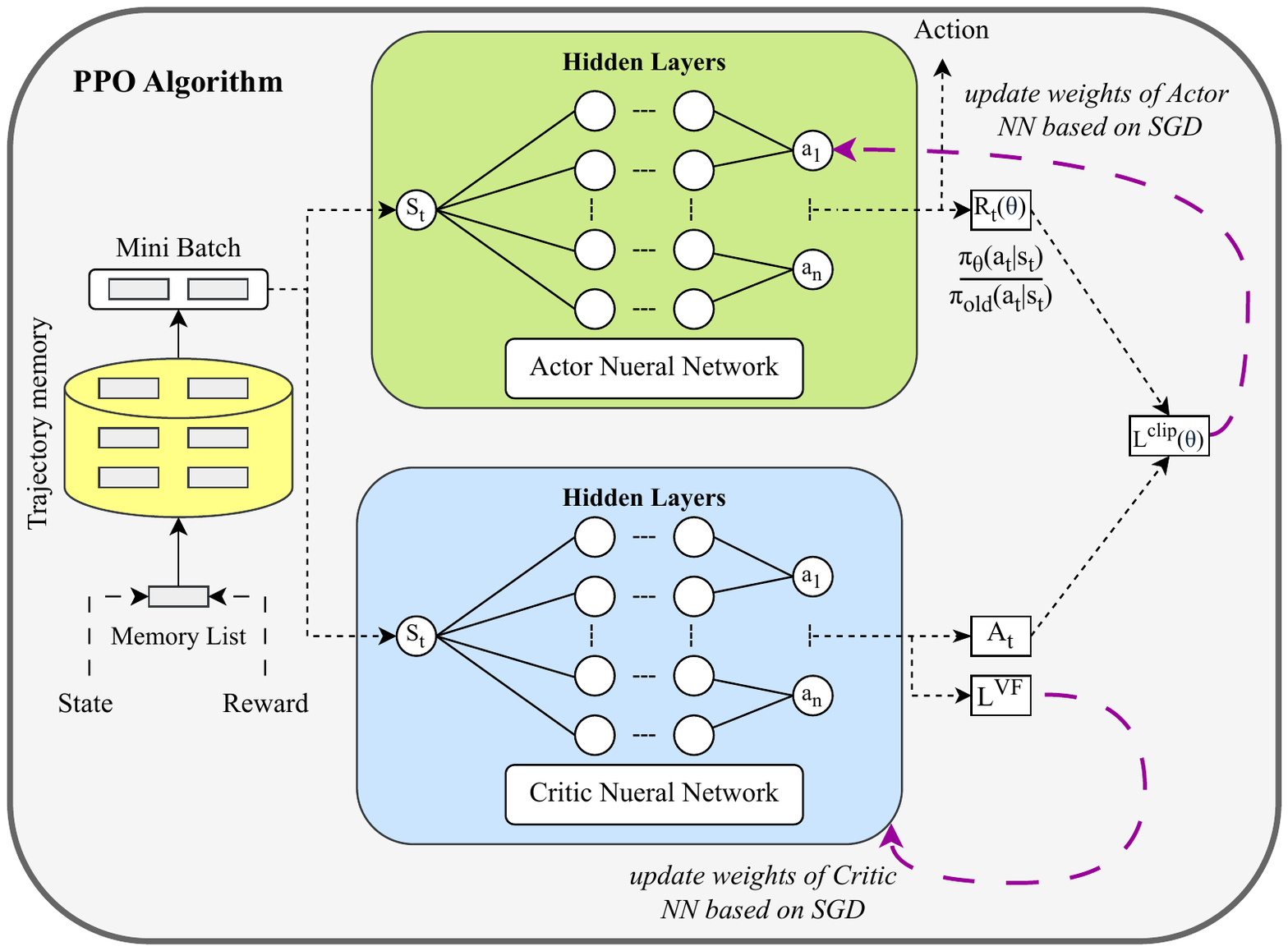}
    \caption{PPO update. The actor and critic networks share a fused state vector composed of visual, LiDAR, PID, and semantic (BLIP) embeddings.}
    \label{fig:ppo_update}
\end{figure}

As illustrated in Fig.~\ref{fig:ppo_update}, At each timestep, the actor-network produces a continuous action vector $(a_1, a_2)$ corresponding to steering and velocity commands, while the critic network estimates the state-value function $V(s_t)$. Generalized Advantage Estimation (GAE) combines rewards and value estimates to produce advantage estimates, which in turn are used to update both the policy (actor) and value function (critic) networks.

\textbf{PPO Optimization Loop:} The main steps of the PPO optimization, as implemented in our pipeline, are outlined in Algorithm ~\ref{alg:ppo_multimodal_full}.

\begin{algorithm}[ht]
	\caption{Pseudocode for the BLIP-FusePPO pipeline}
	\label{alg:ppo_multimodal_full}
	\setlength{\tabcolsep}{3pt} 
	\renewcommand{\arraystretch}{1} %
	\begin{tabular}{p{0.99\linewidth}}
		\textbf{Input:} Multi-modal environment; PPO policy $\pi_\theta$; critic $V_\phi$; reward function $R$; max episode steps $T$. \\[1ex]
		\textbf{Initialization:} Set random seed. Initialize environment, policy parameters $\theta$, critic parameters $\phi$, trajectory memory $\mathcal{M} \leftarrow \emptyset$. \\[1ex]
		\textbf{for} episode $=1$ to $N$ \textbf{do} \\
		\quad Reset environment: \\
		\qquad Obtain initial image $I_0$, LiDAR $L_0$, PID correction $p_0$ \\
		\qquad Generate BLIP token $b_0$ from $I_0$ via VLM \\
		\qquad Set $s_0 = \{I_0, L_0, p_0, b_0\}$ \\
		\quad \textbf{for} $t=0$ to $T$ \textbf{do} \\
		\qquad Sample $a_t \sim \pi_\theta(a_t|s_t)$ \\
		\qquad Execute $a_t$, observe $r_t$, $I_{t+1}$, $L_{t+1}$, $p_{t+1}$ \\
		\qquad Every $K$ steps: generate $b_{t+1}$ from $I_{t+1}$ \\
		\qquad $s_{t+1} = \{I_{t+1}, L_{t+1}, p_{t+1}, b_{t+1}\}$ \\
		\qquad Store $(s_t, a_t, r_t, s_{t+1})$ in $\mathcal{M}$ \\
		\qquad \textbf{if} terminal: \textbf{break} \\
		\quad \textbf{end for} \\
		\quad \textbf{Data Augmentation:} Augment trajectories (e.g., flip images, mirror lidar), add to $\mathcal{M}$ \\
		\quad \textbf{Policy Update:} \\
		\qquad Sample mini-batch from $\mathcal{M}$ \\
		\qquad For each sample: \\
		\qquad \qquad\textbullet~Compute policy ratio: \( R_t(\theta) = \frac{\pi_\theta(a_t | s_t)}{\pi_{\mathrm{old}}(a_t | s_t)} \) \\
		\qquad \qquad\textbullet~Estimate advantage $A_t$ (e.g., GAE) \\
		\qquad \qquad\textbullet~Actor loss: \[ L^{\mathrm{CLIP}}(\theta) = \mathbb{E}_t\left[ \min\Big( R_t(\theta)A_t,\ \mathrm{clip}(R_t(\theta), 1{-}\epsilon, 1{+}\epsilon)A_t \Big) \right] \] \\
		\qquad \qquad\textbullet~Critic loss:\quad $L^{VF}(\phi) = (V_\phi(s_t) - \hat{R}_t)^2$ \\
		\qquad \qquad\textbullet~Update $\phi \leftarrow \phi - \eta_\phi \nabla_\phi L^{VF}$ \\
		\qquad \qquad\textbullet~Update $\theta \leftarrow \theta + \eta_\theta \nabla_\theta L^{\mathrm{CLIP}}$ \\
		\textbf{end for} (episode) \\
		
	\end{tabular}
\end{algorithm}

\section{Simulations and Results}
\label{sec:Results}

\subsection{Simulation Environment}

Experiments were conducted in a custom simulation environment based on Webots, integrated with an OpenAI Gym-compatible interface. This setup enables multimodal perception and realistic control through diverse urban layouts, including straight and curved roads, varying traffic signage, the presence or absence of guardrails, and tunable lighting and weather conditions.

At the start of each episode, the vehicle initial pose is selected from a predefined set in a semi-random cyclic fashion and applied via Supervisor commands. A complete simulator reset is performed at the end of each episode to ensure consistent dynamics. The observation space includes RGB images, LiDAR scans, PID control signals, and semantic tokens generated by a VLM. Semantic features are cached to reduce computational overhead.

\subsection{Training Procedure and Augmentation}

The agent is trained using PPO for 100,000 steps. No fixed episode length is enforced; episodes terminate early if safety thresholds (e.g., lateral deviation or LiDAR proximity) are violated. Each training step includes data augmentation through horizontal image flipping, LiDAR inversion, and PID sign reversal. Semantic tokens remain fixed across augmentations to preserve scene-level consistency.
Reward weights and distance thresholds are summarized in Table~\ref{tab:reward-params}.

\begin{table}[H]
\centering
\caption{Summary of Reward Function Hyperparameters}
\label{tab:reward-params}
\begin{tabular}{|l|c|c|}
\hline
\textbf{Parameter} & \textbf{Value} & \textbf{Note} \\ \hline
\multicolumn{3}{|c|}{\textbf{1) Lane / Centering Terms}} \\ \hline
$w_\mathrm{lane}$, $w_\mathrm{center}$ & 0.3, 0.2 & weights \\
$d_\mathrm{lane}$, $d_\mathrm{clip}$, $d_\mathrm{reset}$ & 100,\ 80,\ 85 & px \\
$k$ & 2.5 & center penalty gain \\
\hline
\multicolumn{3}{|c|}{\textbf{2) Speed Terms}} \\ \hline
$w_\mathrm{speed}$ & 0.2 & speed weight \\
$v_\mathrm{target}$ & 20.0 & km/h \\
\hline
\multicolumn{3}{|c|}{\textbf{3) LiDAR (Obstacle) Terms}} \\ \hline
$w_\mathrm{lidar}$ & 0.3 & lidar weight \\
$d_\mathrm{mid},\, d_\mathrm{low},\, d_\mathrm{crit}$ & 8, 4, 2.8 & m \\
$b_1,\,b_2,\,b_3,\ r_\mathrm{bonus}$ & 5,\ 10,\ 2,\ 5 & penalty/bonus \\
Lidar bonus ranges & $[3,4]\cup[8,10]$ & m \\
$d_\mathrm{fail}$ & 2 & m (terminate) \\
\hline
\multicolumn{3}{|c|}{\textbf{General / Clipping}} \\ \hline
$R_\mathrm{clip}$,\ $r_\mathrm{fail}$ & 1,\ 3 & clip,\ penalty \\
\hline
\end{tabular}
\end{table}

\vspace{14pt}
PPO training settings are detailed in Table~\ref{tab:ppo_hyperparams}. A MultiInputPolicy is used to handle the multimodal input space.
\begin{table}[H]
\centering
\caption{PPO Training Hyperparameters}
\label{tab:ppo_hyperparams}
\begin{tabular}{ll}
\toprule
\textbf{Parameter} & \textbf{Value} \\
\midrule
Policy architecture & MultiInputPolicy \\
Learning rate & $3 \times 10^{-4}$ \\
Rollout steps & $2048$ \\
Batch size & $64$ \\
Epochs per update & $10$ \\
Discount factor $\gamma$ & $0.99$ \\
GAE parameter $\lambda$ & $0.95$ \\
Clipping range $\epsilon$ & $0.2$ \\
Value loss coefficient & $0.5$ \\
Gradient clipping & $0.5$ \\
Execution device & Auto (CPU/GPU) \\
\bottomrule
\end{tabular}
\end{table}

\subsection{Empirical Evaluation and Analysis of Results}

\begin{figure}[t]
    \centering
    \includegraphics[scale=0.4]{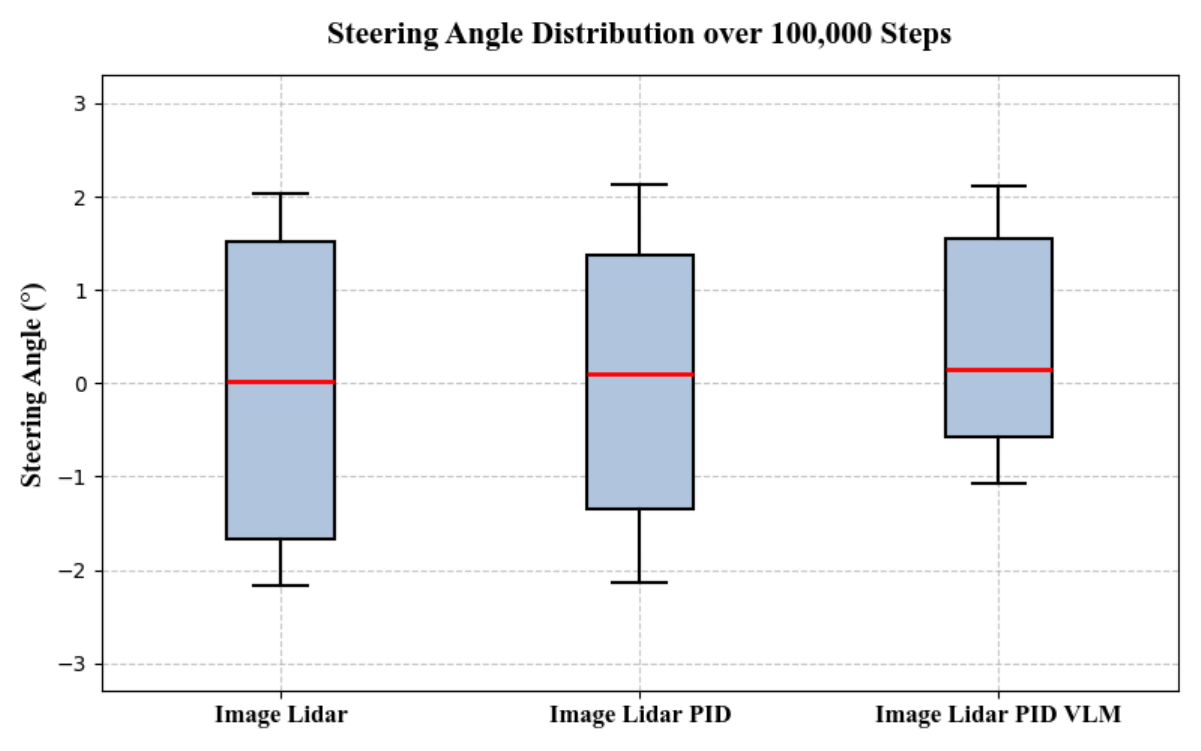}
    \caption{Steering angle distributions for various agent configurations. The PID+VLM-enhanced agent exhibits tightly clustered, low-variance steering angles near zero (straight driving), reflecting more stable policy behavior.}
    \label{fig:steering_boxplot}
\end{figure}

\begin{figure*}[!t]
    \centering
    \includegraphics[width=0.97\linewidth]{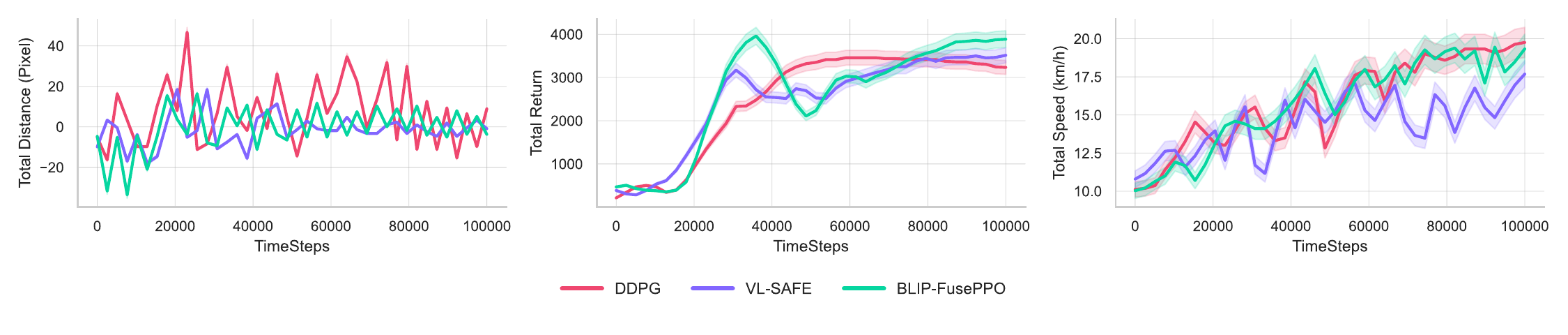}
    \caption{Comparative training curves of the DDPG \cite{perez2022deep}, VL-SAFE \cite{qu2025vl}, and BLIP-FusePPO approaches: lateral deviation, cumulative reward, and average speed.}
    \label{fig:main_results}
\end{figure*}

The stability and effectiveness of the training process are illustrated in Figs.~\ref{fig:main_results}--\ref{fig:reward_func}. Key findings, ablations, and analytical insights based on the simulation data are presented as follows:

\subsubsection{Comparison of State Enrichment Approaches (Figs.~\ref{fig:main_results}, \ref{fig:steering_boxplot})}

\textit{Lateral Deviation:}  
As demonstrated in the left plot of Fig.~\ref{fig:main_results}, agents equipped with both PID and VLM semantic features (BLIP-FusePPO) achieve the lowest mean tracking error and minimal oscillation. These results highlight the benefit of multi-modal state enrichment, enabling the agent to maintain accurate and stable LK even under noisy or adversarial conditions.

\textit{Cumulative Reward:}  
In the middle plot, Fig.~\ref{fig:main_results} shows faster convergence and higher cumulative reward for enriched agents. Inclusion of PID and semantic tokens results in increased learning stability and superior control quality, as opposed to slower, oscillatory progress in the baseline models.

\textit{Speed Tracking:}  
The right plot of Fig.~\ref{fig:main_results} shows that the proposed agent maintains speeds closer to the desired target with reduced abruptness. Baseline methods exhibit unstable acceleration and more abrupt throttle/braking actions.

\textit{Steering Angle Distribution:}  
As shown in Fig.~\ref{fig:steering_boxplot}, adding PID and VLM state signals results in a tighter steering distribution, with the median angle close to zero and reduced variance. This indicates fewer abrupt corrections and more confident, stable control behavior.

\begin{figure*}[!t]
    \centering
    \includegraphics[width=0.9\linewidth]{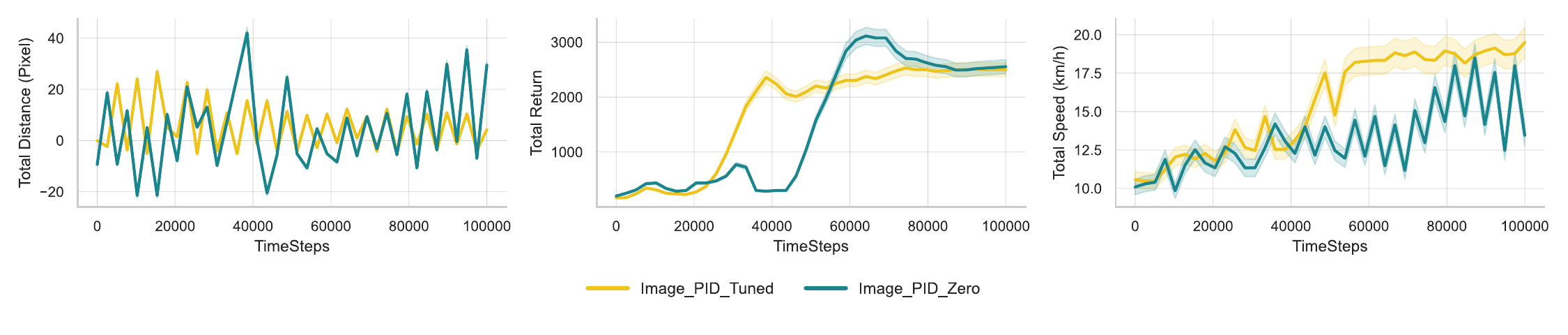}
    \caption{Effect of PID signal: Comparing PID-tuned and non-PID configurations. Properly tuned PID noticeably improves stability, accelerates reward convergence, and produces smoother agent behavior. Without control grounding, higher variance and sudden behavioral collapse are observed.}
    \label{fig:pid_ablation}
\end{figure*}

\begin{figure*}[!t]
    \centering
    \includegraphics[width=0.9\linewidth]{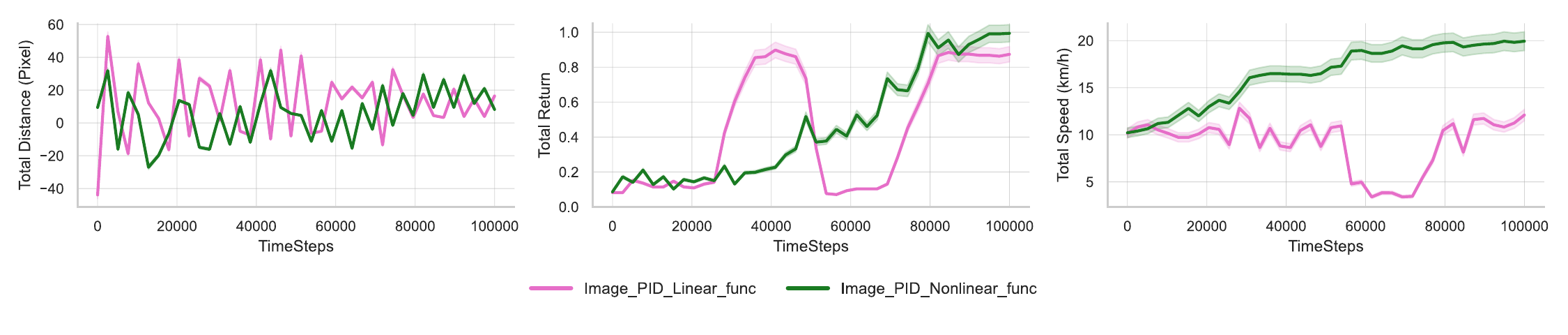}
    \caption{Nonlinearity in reward shaping: Nonlinear and weighted reward formulations lead to lower lateral deviation, smoother reward growth, and better-controlled speeds compared to linear versions, which often produce divergence or instability.}
    \label{fig:reward_func}
\end{figure*}

\subsubsection{Ablation Studies: PID Signal and Reward Design (Figs.~\ref{fig:pid_ablation},~\ref{fig:reward_func})}

\textit{Effect of PID Signal:}  
Fig.~\ref{fig:pid_ablation} directly compares agents with and without PID feedback. Injecting a well-tuned PID signal consistently enhances stability, accelerates reward convergence, and reduces erratic or collapsed behaviors. The absence of explicit control signals results in higher variance and abrupt policy failures.

\textit{Reward Function Nonlinearity:}  
Results in Fig.~\ref{fig:reward_func} confirm that the nonlinear, weighted reward formulation yields lower lateral deviations, smoother reward increases, and more reliable speed regulation. A purely linear reward setup causes higher instability and can lead to early divergence.

\subsection{Quantitative Validation of LK Accuracy}
To rigorously assess lateral control accuracy across various episodes, we employed metrics grounded in regression analysis, including Root Mean Square Error (RMSE), normalized RMSE (nRMSE), and the standard deviation (std) of the vehicle lateral position relative to the lane centerline.

At each time step $i$, the lateral offset measured in pixels ($y_i$) is converted to meters ($d_i$) using a predefined scale factor derived from image-to-world correspondence:
\[ \text{Scale}_{px \to m} = \frac{5}{235} = 0.02128 \]
\begin{equation*}
d_i = y_i \times 0.02128
\end{equation*}
where $d_i$ denotes the lateral deviation in meters. The aggregate performance over $N$ frames is then evaluated via:
\begin{equation*}
\mathrm{RMSE} = \sqrt{ \frac{1}{N} \sum_{i=1}^{N} (d_i)^2 }
\end{equation*}
\begin{equation*}
\mathrm{Std} = \sqrt{ \frac{1}{N} \sum_{i=1}^{N} (d_i - \bar{d})^2 }
\end{equation*}
Here, $\bar{d}$ represents the mean lateral deviation across all frames.

The normalized RMSE is calculated as:
\begin{equation*}
\mathrm{nRMSE} = \frac{\mathrm{RMSE}}{5}
\end{equation*}
Where $5~\textrm{m}$ denoting the real-world lane width.

To improve statistical robustness, all metrics were averaged over 100 independent test drives, each featuring randomized initial positions and lane configurations. Table~\ref{tab:lane_quantitative} shows the quantitative performance of three LK controllers that were tested 100 times by different people. The proposed method, BLIP-FusePPO, has the lowest RMSE of 0.110,m, which means it is more accurate than DDPG~\cite{perez2022deep} (0.242,m) and VL-SAFE~\cite{qu2025vl} (0.198,m). Moreover, the standard deviation of lateral deviations for BLIP-FusePPO is now only 0.055 m, which shows that it is more stable and consistent in a wider range of driving conditions. nRMSE, which is based on a lane width of 5 m, shows even more performance improvements. BLIP-FusePPO got 0.0220, while VL-SAFE got 0.0396, and DDPG got 0.0484. The results demonstrate the effectiveness of the proposed multimodal fusion approach and the integration of vision-language representations within the Proximal Policy Optimization (PPO) framework.

\begin{table}[ht]

\caption{Quantitative LK results over 100 trials}
\begin{tabular}{lccc}
\toprule
\textbf{Method} & \textbf{RMSE (m)} & \textbf{Std. Dev. (m)} & \textbf{nRMSE} \\
\midrule
DDPG~\cite{perez2022deep} & 0.242 & 0.121 & 0.0484 \\
VL-SAFE~\cite{qu2025vl}    & 0.198 & 0.099 & 0.0396 \\
\textbf{BLIP-FusePPO (Proposed)} & \textbf{0.110} & \textbf{0.055} & \textbf{0.0220} \\
\bottomrule
\end{tabular}
\label{tab:lane_quantitative}
\end{table}

\section{Discussion and Conclusion}
\label{sec:Conclusion}

This work presents BLIP-FusePPO, a novel multimodal RL method for autonomous LK. By combining semantic embeddings from a VLM with geometric state information, LiDAR sensor readings, and PID-based control feedback, our method provides a dense, interpretable state representation. This fusion allows the agent to make context-aware decisions, bolstering its capacity to drive in challenging and dynamic environments more more accurate and robust.

The suggested hybrid reward function, fusing semantic alignment, lane compliance, obstacle avoidance, and velocity control, allows for effective and generalizable policy learning. In contrast to previous approaches that use VLMs exclusively for reward shaping, BLIP-FusePPO incorporates semantic features directly into the observation space. This construction provides persistent semantic awareness while dramatically lowering computational overhead at inference time, rendering it highly suitable for real-time deployment.

Comprehensive simulation show that BLIP-FusePPO surpasses state-of-the-art vision-based and multimodal reinforcement learning baselines with up to 54.5\% less RMSE than DDPG and 44.4\% better performance than VL-SAFE on varied and complex driving maneuvers. These findings highlight the effectiveness of our multimodal fusion approach and its potential for improving LK stability and adaptiveness.

\bibliographystyle{IEEEtran}
\bibliography{my_bib}
\vspace{-48pt}  

\begin{IEEEbiography}[{\includegraphics[width=1in, height=1.25in, clip, keepaspectratio]{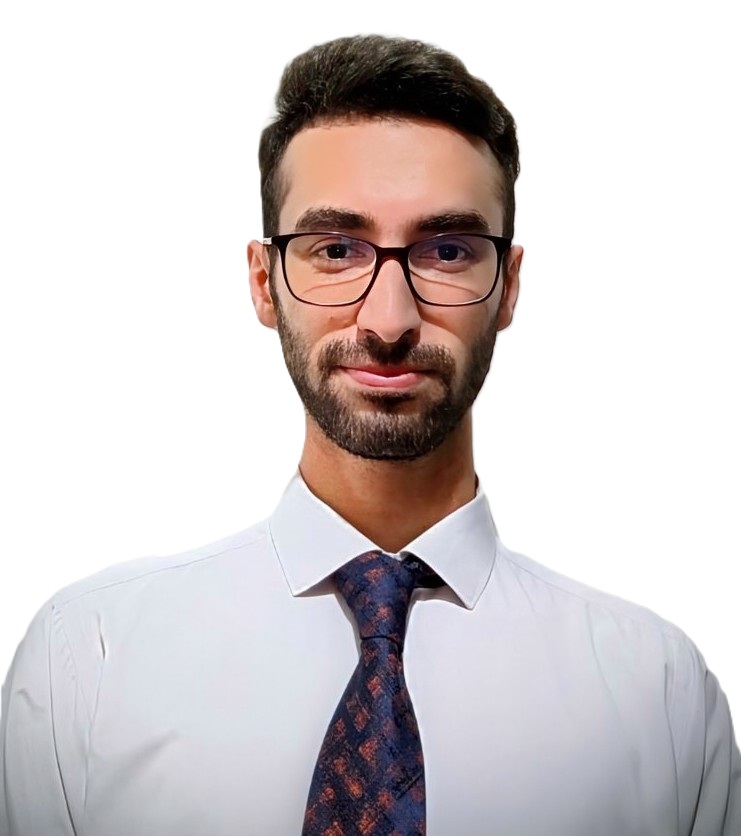}}]{Seyed Ahmad Hosseini Miangoleh}
is currently pursuing the B.S. degree in electrical engineering (control) at Amirkabir University of Technology, Tehran, Iran. His research interests include control systems, robotics, machine learning, deep learning, reinforcement learning, and computer vision.
\end{IEEEbiography}
\vspace{-48pt}  

\begin{IEEEbiography}[{\includegraphics[width=1in, height=1.25in, clip, keepaspectratio]{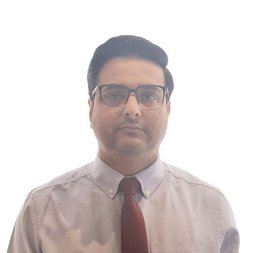}}]{Amin Jalal Aghdasian}
received the B.S. degree in electrical engineering from Tabriz University, Tabriz, Iran, in 2018, and the M.S. degree in mechatronics engineering from Amirkabir University of Technology, Tehran, Iran, in 2023. His research interests include intelligent automotive systems, robust control, machine learning, neural networks, robotics, and reinforcement learning.
\end{IEEEbiography}
\vspace{-48pt}  

\begin{IEEEbiography}[{\includegraphics[width=1in, height=1.25in, clip, keepaspectratio]{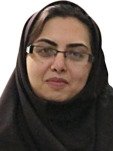}}]{Farzaneh Abdollahi}
(Senior Member, IEEE) received the B.Sc. degree in electrical engineering from the Isfahan University of Technology, Isfahan, Iran, in 1999, the M.Sc. degree in electrical engineering from the Amirkabir University of Technology, Tehran, Iran, in 2003, and the Ph.D. degree in electrical engineering from Concordia University, Montreal, QC, Canada, in 2008. She is currently an Associate Professor with the Amirkabir University of Technology and an adjunct Professor with Carleton University. Her research interests include intelligent control, robotics, control of nonlinear systems, control of multiagent networks, and robust and switching control.
\end{IEEEbiography}

\end{document}